\documentclass[runningheads]{llncs}
\usepackage{graphicx}

\usepackage{tikz}
\usepackage{comment}
\usepackage{amsmath,amssymb} 
\usepackage{color, colortbl} 
\usepackage{subcaption}
\usepackage{pifont}

\usepackage[accsupp]{axessibility}  

\begin{document}
\pagestyle{headings}
\mainmatter
\def\ECCVSubNumber{361}  

\newcommand{\cmark}{\text{\ding{51}}}%
\newcommand{\xmark}{\text{\ding{55}}}%
\newcommand{\R}[1]{\textcolor[rgb]{1.00,0.00,0.00}{#1}}
\newcommand{\B}[1]{\textcolor[rgb]{0.00,0.00,1.00}{#1}}
\definecolor{Gray}{gray}{0.92}

\title{Overexposure Mask Fusion: Generalizable Reverse ISP Multi-Step Refinement} 

\titlerunning{Overexposure Mask Fusion for Reverse ISP Refinement} 
%

\author{Jinha Kim \inst{1, 2} \and
Jun Jiang \inst{2} \and
Jinwei Gu\inst{2}}

%
\authorrunning{J. Kim et al.}
%

\institute{MIT, Cambridge, MA \\ 
\email{jinhakim@mit.edu} 
\and 
SenseBrain Technology, San Jose, CA \\ 
\email{\{jinhakim,jiangjun,gujinwei\}@sensebrain.site}} 

\maketitle

\begin{abstract}
With the advent of deep learning methods replacing the ISP in transforming sensor RAW readings into RGB images, numerous methodologies solidified into real-life applications. Equally potent is the task of inverting this process which will have applications in enhancing computational photography tasks that are conducted in the RAW domain, addressing lack of available RAW data while reaping from the benefits of performing tasks directly on sensor readings. This paper’s proposed methodology is a state-of-the-art solution to the task of RAW reconstruction, and the multi-step refinement process integrating an overexposure mask is novel in three ways: instead of from RGB to bayer, the pipeline trains from RGB to demosaiced RAW allowing use of perceptual loss functions; the multi-step processes has greatly enhanced the performance of the baseline U-Net from start to end; the pipeline is a generalizable process of refinement that can enhance other high performance methodologies that support end-to-end learning. 
\keywords{ISP, Reversed ISP, Demosaiced RAW, Multi-Step Refinement, Overexposure Mask}
\end{abstract}

\section{Introduction} 

Image signal processor (ISP) denotes a collection of operations integrated in today’s digital cameras that maps camera sensor readings into visually pleasing RGB images. A popular area of research that has been explored in relation to the ISP is the task of mapping from RAW data to RGB images with the use of deep learning-based methodologies. With various applications such as in mobile cameras which have small sensors and other limitations in hardware, various  methodologies \cite{ignatov_van_gool_timofte_2020,schwartz_giryes,xing_qian_chen_2021} have been developed to address this task.  

A problem that also relates to the ISP, which has equally potent applications as the task of mapping RAW data to RGB images, is the reversed task of mapping from RGB images to RAW data, which is a novel problem in low-level computer vision. Unlike RGB images, RAW data holds a linear relationship with scene irradiance, which has led to improved performance in various computer vision tasks. Numerous works have addressed the task of RAW reconstruction with various methodologies with solutions ranging from utilizing canonical steps approximated by invertible functions \cite{brooks_mildenhall_xue_chen_sharlet_barron_2019}, mapping RAW data to CIE-XYZ space from sRGB images \cite{afifi_abdelhamed_abuolaim_punnappurath_brown_2021}, a novel modular and differentiable ISP model with interpretable parameters that is capable of end-to-end learning \cite{conde_mcdonagh_maggioni_leonardis_pérez-pellitero_2022} among many approaches \cite{afifi_abdelhamed_abuolaim_punnappurath_brown_2021,brooks_mildenhall_xue_chen_sharlet_barron_2019,conde_mcdonagh_maggioni_leonardis_pérez-pellitero_2022,punnappurath_brown_2020,xing_qian_chen_2021,zamir_arora_khan_hayat_khan_yang_shao_2020}. With these inherent advantages that RAW data holds, the task of reconstructing RAW data from RGB images has become exceedingly relevant, especially with the lack of availability of RAW data due to factors such as memory-related concerns or data storage processes that discard the RAW. 

However, the task of RAW data reconstruction remains a novel area of research with  complexities and limitations that are yet to be fully addressed. For instance, as noted by Conde et al. \cite{conde_mcdonagh_maggioni_leonardis_pérez-pellitero_2022}, approximations using inverse functions for real-world ISPs show degradation in performance when a large portion of the RGB images are close to overexposure. Our proposed methodology using overexposure mask fusion is a novel portion of our pipeline that specifically addresses this issue by mapping overexposed and non-overexposed pixels separately and fusing them together using an overexposure mask. 

Among various AIM challenges with different research problems \cite{ignatov2022isp}, for the AIM Reversed ISP Challenge \cite{conde2022aim} where competing teams were given the task of reconstructing RAW data from RGB images, our methodology is a top solution, and therefore, evaluated as a state-of-the-art solution to the novel inverse problem. By mapping from RGB to demosaiced RAW by generating a demosaiced RAW from the groundtruth bayer using Demosaic Net \cite{gharbi_chaurasia_paris_durand_2016}, we allow the use of perceptual losses. With our novel overexposure mask fusion methodology, our pipeline addresses the issue of overexposed pixels as mentioned by Conde et al. \cite{conde_mcdonagh_maggioni_leonardis_pérez-pellitero_2022}. It is most notable that the pipeline led to significant enhancement in fidelity measures while keeping all neural networks within our pipeline as the U-Net \cite{DBLP:journals/corr/RonnebergerFB15}. It is further notable that our methodology can incorporate other proposed state-of-the-art solutions involving end-to-end learning after slight modifications to map from RGB images to demosaiced RAW images. For instance, the model proposed by Conde et al. \cite{conde_mcdonagh_maggioni_leonardis_pérez-pellitero_2022} can be integrated with our refinement pipeline by making small modifications such as removing the final mosaic step and generating demosaiced RAW groundtruth images for training in order to use perceptual loss. We propose, to the best of our knowledge, the first generalizable, multi-step refinement process for enhanced performance of other reversed ISPs while addressing the issue of overexposure.

\section{Related Works} 

Works such as \cite{ignatov_van_gool_timofte_2020,schwartz_giryes,xing_qian_chen_2021} have addressed the task of mapping from RAW data to RGB images, modeling the camera ISP. Schwartz et al. \cite{schwartz_giryes} proposes a full end-to-end deep learning model of the ISP, which has demonstrated to be capable of generating visually compelling RGB images from RAW data. Ignatov et al. \cite{ignatov_van_gool_timofte_2020} proposes another end-to-end deep learning solution with the use of a novel PyNET CNN architecture and Xing et al. \cite{xing_qian_chen_2021} designed an invertible ISP that is capable of generating visually pleasing RGB images from RAW data as well as RAW reconstruction. Another work is the CycleISP \cite{zamir_arora_khan_hayat_khan_yang_shao_2020} which models the ISP both in the forward and reverse directions. 

There have also been various works addressing the task of RAW reconstruction from RGB images \cite{afifi_abdelhamed_abuolaim_punnappurath_brown_2021,brooks_mildenhall_xue_chen_sharlet_barron_2019,conde_mcdonagh_maggioni_leonardis_pérez-pellitero_2022,punnappurath_brown_2020,xing_qian_chen_2021,zamir_arora_khan_hayat_khan_yang_shao_2020}. Brooks et al. \cite{brooks_mildenhall_xue_chen_sharlet_barron_2019} proposes an unprocessing technique for RAW reconstruction by inverting the ISP pipeline with five canonical steps that are approximated by invertible functions while CIE-XYZ Net \cite{afifi_abdelhamed_abuolaim_punnappurath_brown_2021} recovers the RAW data to the CIE-XYZ space from sRGB images. Conde et al. \cite{conde_mcdonagh_maggioni_leonardis_pérez-pellitero_2022} proposed a novel modular and differentiable ISP model with interpretable parameters and canonical camera operations that is capable of end-to-end learning of parameter representations. Punnappurath et al. \cite{punnappurath_brown_2020} proposed modifications to loss used for training neural network-based compression architectures to account for both sRGB image fidelity and RAW reconstructions errors while modeling sRGB-RAW mapping with the use of locally differentiable 3D lookup tables.  Previously mentioned for the task of mapping from RAW to RGB, CycleISP \cite{zamir_arora_khan_hayat_khan_yang_shao_2020} and the invertible ISP model proposed by Xing et al. \cite{xing_qian_chen_2021} are also capable of RAW reconstruction. Several works offering solutions to the task of RAW reconstruction after integration of their approaches of RAW reconstruction have noted improvements in performance for RAW image denoising \cite{brooks_mildenhall_xue_chen_sharlet_barron_2019,conde_mcdonagh_maggioni_leonardis_pérez-pellitero_2022,zamir_arora_khan_hayat_khan_yang_shao_2020} which suggests further applications of RAW reconstruction. 

In order to evaluate performance of different solutions on the task of RAW reconstruction for the AIM Reversed ISP Challenge \cite{conde2022aim}, two datasets were used for training which are the Samsung S7 dataset \cite{schwartz_giryes} and ETH Huawei P20 Pro dataset dataset \cite{ignatov_van_gool_timofte_2020}. The Samsung S7 dataset \cite{schwartz_giryes} consists of $110$ scenes of $3024 \times 4032$ resolution as JPEG images captured with a Samsung S7 rear camera where original RAW images were saved as well. The ETH Huawei P20 Pro dataset \cite{ignatov_van_gool_timofte_2020} is a large-scale dataset consisting of $20$ thousand photos collected using a Huawei P20 smartphone for capturing RAW images and the RGB images obtained with Huawei’s built-in ISP (12.3 MP Sony Exmor IMX380). For both tracks, participants were evaluated on fidelity measures, PSNR and SSIM, and were also tested for generizability and robustness of proposed methods.  

\section{Methodology} 

\subsection{Network Architecture} 

\begin{figure}[!ht]
    \centering
    \begin{minipage}{1\textwidth} 
    \includegraphics[width=1\textwidth]{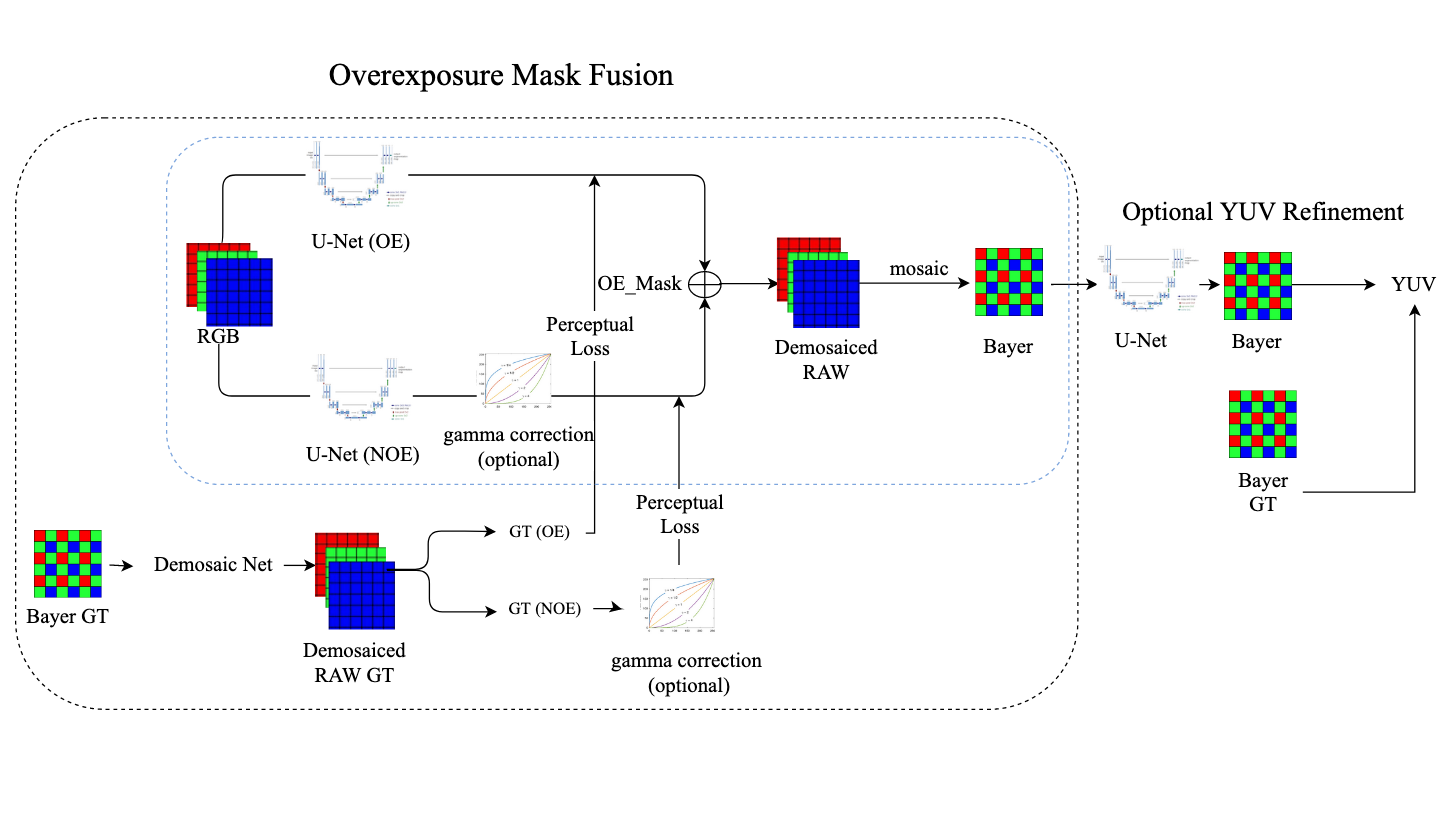} 
    \end{minipage} 

    \caption{The architecture of the proposed pipeline. Neural networks were all unified to be U-Nets for objective comparisons. U-Net (overexposure or OE) and U-Net (non-overexposure or NOE) both take in input RGB images and outputs demosaiced RAW. However, non-overexposed pixels are set to $0$ for U-Net (OE), GT(OE) and overexposed pixels are set to $0$ for U-Net (NOE), GT(NOE). It is notable that the U-Nets can easily be replaced with another methodology that supports end-to-end learning for mapping from RGB to demosaiced RAW.} 
    \label{network_architecture}
\end{figure}

The schematic representation of the overall pipeline is outlined in  Fig.~\ref{network_architecture}. The general structure of pipeline consists of unprocessing the input RGB image to its original demosaiced RAW, after which a simple mosaic is performed to recover to bayer. For training, the pipeline involves generating new groundtruth RGB images by passing the groundtruth bayer through a pretrained Demosaic Net \cite{gharbi_chaurasia_paris_durand_2016} in order to reconstruct the demosaiced RAW. Notably, unlike methodologies that map directly from RGB to bayer, the proposed pipeline maps initially from RGB to demosaiced RAW, which enables the use of perceptual loss functions \cite{zhang_isola_efros_shechtman_wang_2018}. 

\begin{figure}[!ht]
    \centering
    \begin{minipage}{1\textwidth} 
    \includegraphics[width=1\textwidth]{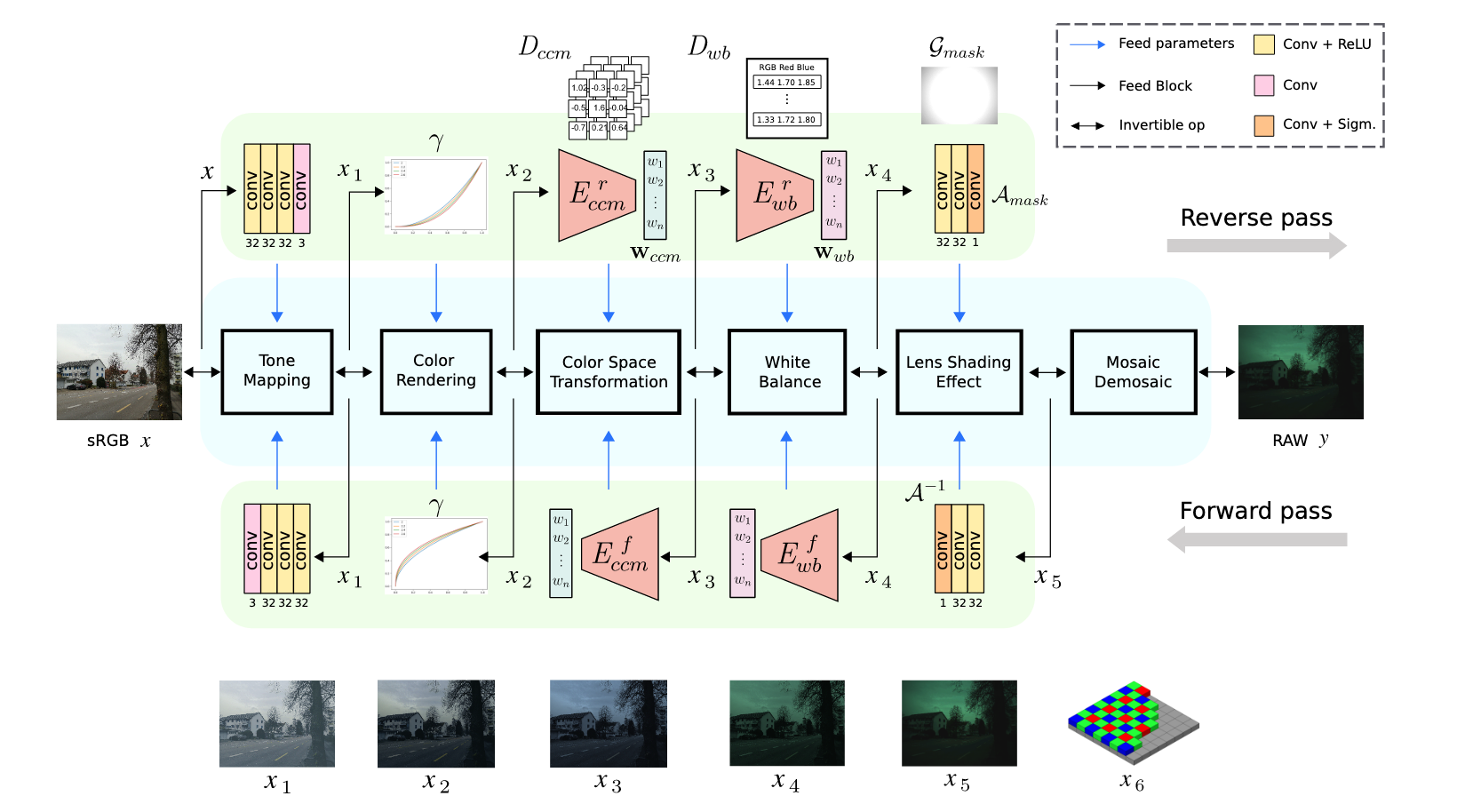} 
    \end{minipage}

    \caption{The figure displays a state-of-the-art model proposed by Conde et al. \cite{conde_mcdonagh_maggioni_leonardis_pérez-pellitero_2022} that maps from sRGB to RAW. Different methodologies supporting end-to-end learning can be integrated as replacements for our U-Nets \cite{DBLP:journals/corr/RonnebergerFB15} after slight modifications to map from sRGB to demosaiced RAW.}  
    \label{network_architecture_comparison}
\end{figure}

A binary overexposure mask is constructed by computing the illuminance for each pixel of the input RGB image. Based on a certain threshold, pixels are then marked as overexposed or non-overexposed. As displayed in Fig.~\ref{network_architecture}, there are two separate U-Nets \cite{DBLP:journals/corr/RonnebergerFB15}, U-Net (overexposure or OE) and U-Net (non-overexposure or NOE), which takes in the same input RGB image, but are trained to account respectively for overexposed and non-overexposed pixels. The U-Net (OE) and U-Net (NOE) both have $23$ convolutional layers with the slight modification of the additional channel to include the overexposure mask. It is noteworthy that the two U-Nets can be replaced by any other high performance methodology that can support end-to-end learning for mapping RGB to demosaiced RAW as displayed in Fig.\ref{network_architecture_comparison}. For purposes of ablation, the U-Net \cite{DBLP:journals/corr/RonnebergerFB15} was the only neural network model used for inference throughout the pipeline. 

The two neural networks each output a demosaiced RAW image after which loss for U-Net (overexposure or OE) and U-Net (non-overexposure or NOE) are computed separately with the use of the overexposure mask. By generating two new groundtruth images, GT (OE) and GT (NOE), each groundtruth image being created by setting the pixels that are not respectively accounted for to $0$, for instance, non-overexposed pixels for GT (OE) to $0$. This operation is similarly performed on the outputs of the U-Nets such that U-Net (OE) and GT (OE) have value $0$ for non-overexposed pixels, and conversely for U-Net (NOE) and GT (NOE). Additionally, for U-Net (NOE), gamma correction can be applied on non-overexposed pixels before computing loss to account for low-light demosaiced RAW images. 

Although the training process requires the use of Demosaic Net \cite{gharbi_chaurasia_paris_durand_2016} to generate new groundtruth images, inference, as marked by the blue container in Fig.\ref{network_architecture}, involves passing RGB input into U-Net (OE) and U-Net (NOE), after which the two images are simply blended together with the use of the overexposure mask. Inference is then completed by taking the demosaiced RAW output and mosaicing to convert to  bayer.\footnotemark[1] \footnotetext[1]{Starting code provided by the AIM Reversed ISP Challenge Organizers, which is available at https://github.com/mv-lab/AISP, was used within our training and inference code.} The pipeline includes an optional YUV refinement step that computes loss between bayer in YUV space. It is most notable that the pipeline's multi-step refinement process can be applied to other reversed ISPs that support end-to-end learning with slight modifications to map from RGB to demosaiced RAW. Furthermore, as conducted for the ETH Huawei P20 Pro dataset \cite{ignatov_van_gool_timofte_2020}, the pipeline can be modified to use a single U-Net with an additional channel to include the overexposure mask, reducing the total number of model parameters while maintaining high performance in fidelity measures. 

\subsection{Reconstructing Demosaiced RAW} 
\begin{figure}[!ht]
    \centering
    \begin{minipage}{0.8\textwidth} 
    \includegraphics[width=1\textwidth]{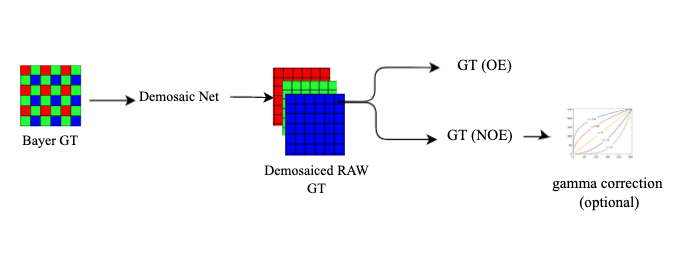} 
    \end{minipage} 

    \caption{This figure illustrates the process of reconstructing two separate groundtruth images from the demosaiced RAW created by passing the bayer groundtruth into a pre-trained Demosaic Net \cite{gharbi_chaurasia_paris_durand_2016}. The two groundtruth images are used to separately compute loss for overexposed and non-overexposed pixels.} 
    \label{reconstruct_demosaiced_RAW}
\end{figure}

The bayer groundtruth is passed into a pretrained demosaic Net and separated into two different groundtruths, for U-Net (OE) and U-Net (NOE) as displayed in Fig.~\ref{reconstruct_demosaiced_RAW}. Gamma correction can be applied for non-overexposed pixels before computing loss with a parameter $\gamma$. For the ETH Huawei P20 Pro dataset \cite{ignatov_van_gool_timofte_2020}, the value of $\gamma = \dfrac{1}{3.6}$ was applied on the normalized RGB values before computing loss for non-overexposed pixels. 

Note that even if gamma correction is applied on overexposed pixels, which have high illuminance, there will be minimal changes by nature of gamma correction. For the model used to train and run inference on ETH Huawei P20 Pro dataset \cite{ignatov_van_gool_timofte_2020}, gamma correction was applied on all pixels as a whole before computing loss, and for the complete pipeline that utilizes two separate neural networks, gamma correction can be applied on non-overexposed pixels for $\gamma \in [0,1]$  

\begin{align}
    L_{non OE} & = L(I_{recon}^{\gamma}, I_{GT}^{\gamma}).    
\end{align} 

By applying gamma correction before computing loss, there was enhancement in performance in perceptual quality of demosaiced RAW, as demosaiced RAW images tend to be low-light, which can be accounted for by increasing the weighting of smaller pixel values within the loss function. 

\subsection{YUV Overexposure Mask} 
Given the input RGB, the YUV overexposure mask is computed by converting RGB pixel values to YUV. Since the  overexposure mask utilizes only illuminance, only Y values are stored from the matrix multiplication below 

\begin{align}
      \begin{bmatrix}
        Y \\ 
        U \\ 
        V 
      \end{bmatrix} 
      = 
      \begin{bmatrix}
         0.299 & 0.587 & 0.114 \\ 
        -0.14713 & -0.28886 & 0.436 \\ 
        0.615 & -0.51499 & -0.10001
     \end{bmatrix}
     \begin{bmatrix}
         R \\ 
         G \\ 
         B 
     \end{bmatrix}.
\end{align}

Pixels with Y $\geq 0.978$ were marked as $1$ and $0$ otherwise. For input RGB patches with size $504 \times 504$, the overexposure mask is binary and has dimensions $504 \times 504$ as well. Note that alternatively the overexposure mask can instead with the threshold of $\max(R, G, B) \geq 0.99$, however, as the performance in fidelity dropped for images with specifically white saturated pixels, the pipeline constructs the mask using illuminance. 

\subsection{Overexposure Mask Fusion for Inference} 
\begin{figure}[!ht]
    \centering
    \begin{minipage}{0.8\textwidth} 
    \centering
    \includegraphics[width=1\textwidth]{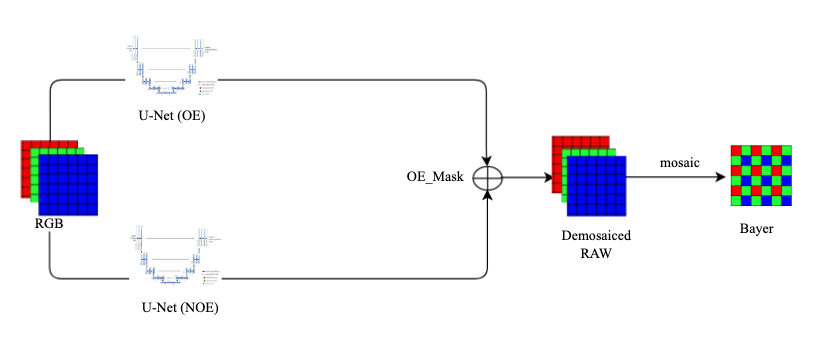} 
    \end{minipage} 

    \caption{The figure displays the fusion of the two output images using the overexposure mask. U-Net (OE) and U-Net (NOE) reconstruct images of same dimensions, and fusion is done by storing the RGB values for overexposed pixels from U-Net (OE) and non-overexposed pixels from U-Net (NOE). This process reconstructs the demosaiced RAW which is then mosaiced to bayer.} 
    \label{overexposure_mask_fusion}
\end{figure}

It is noteworthy that the pipeline does not require use of constructing new groundtruth images for inference, which only requires fusion of the outputs of U-Net (OE), U-Net (NOE). As shown in Fig.~\ref{overexposure_mask_fusion}, the two reconstructed images are blended together using the overexposure mask. 

For U-Net (OE), all non-overexposed RGB pixel values of the output image are set to $0$ while conversely for the output image of U-Net (NOE). Then, by simple addition of the two images, the two images are fused together, resulting in better accuracy in both overexposed and  non-overexposed pixels. As the neural networks were trained to map RGB to its original demosaiced RAW from the groundtruth bayer, the pipeline reconstructs bayer by simply mosaicing of the demosaiced RAW output. 

\subsection{Perceptual Loss Functions and Training Details} 
There is greater availablity of literature on perceptual loss functions in comparison to loss functions between RAW bayer. For training and inference on the Samsung S7 dataset \cite{schwartz_giryes}, different perceptual loss functions were utilized for overexposed pixels and non-overexposed pixels. 

For overexposed pixels, the loss function MS-SSIM-L1 \cite{zhao_gallo_frosio_kautz_2017} was used and for non-overexposed pixels, 

\begin{align}
      L_{NOE} = L_{LPIPS} + L_{2} \cdot 0.05 + L_{MSSSIM-L1} \cdot 0.75, 
\end{align}

LPIPS with AlexNet \cite{zhang_isola_efros_shechtman_wang_2018} and MS-SSIM-L1 \cite{zhao_gallo_frosio_kautz_2017} and L2 was combined in order to improve perceptual quality. A small combination of L2 is taken for fidelity improvement and prevention of notable color deviations. After $200$ epochs with batch size of $6$ with the Adam optimizer \cite{adamoptimizer} of learning rate $10^{-4}$ and decay of $10^{-6}$, both overexposed and non-overexposed pixels were trained again only with L2 loss for an additional $20$ epochs with learning rate of $10^{-5}$. Inference on on Nvidia's Tesla-V100 GPU for patches of $504 \times 504 \times 3$ takes $0.05$ seconds on average per patch.

For training and inference on the ETH Huawei P20 Pro dataset \cite{ignatov_van_gool_timofte_2020} where one instead of two U-Nets was used, the loss function with $\gamma = \dfrac{1}{3.6}$ was computed as 

\begin{align}
      L = L_{LPIPS}(I_{in}^\gamma, I_{gt}^\gamma) + L_2(I_{in}^\gamma, I_{gt}^\gamma) \cdot 0.05 + L_{MSSSIM-L1}(I_{in}^\gamma, I_{gt}^\gamma) \cdot 0.75  
\end{align}

where gamma correction was applied to both overexposed and non-overexposed pixels, as raising overexposed pixel values to $\gamma$ results in minimal changes by nature of gamma correction. Similarly, the Adam optimizer \cite{adamoptimizer} was used with an initial learning rate of $10^{-4}$ with batch size of $6$ and the weight decay was set as $10^{-6}$ and trained for $230$ epochs. Inference with GPU on patches of $496 \times 496 \times 3$ takes on average, $0.00465$ seconds per patch. 

\subsection{Bayer to Bayer Optional Refinement} 

The optional refinement step of the pipeline is refinement from bayer to bayer by computing L1 or L2 loss after mapping each of the bayer into YUV space as displayed in Fig.~\ref{network_architecture} For an input RGB image with dimensions $504 \times 504 \times 3$ to bayer with dimensions $252 \times 252 \times 4$, the refinement step consists of averaging the green channels of the bayer and performing matrix multiplication in order to map to YUV space. By taking L1 or L2 loss on YUV, the reconstructed bayer displayed better fidelity scores than directly computing L1 or L2 on bayer, as will be further detailed in ablation studies. 

\section{Experiments} 

\subsection{Quantitative and Qualitative Evaluations} 

\begin{table}[!ht] 
    \centering
    \caption{AIM Reversed ISP  Challenge Benchmark \cite{conde2022aim}. Teams are ranked based on their performance on \underline{Test1} and \underline{Test2}, an internal test set to evaluate the generalization capabilities and robustness of the proposed solutions. The methods (*) have trained using extra data from~\cite{schwartz_giryes}, and therefore only results on the internal datasets are relevant. CycleISP~\cite{zamir_arora_khan_hayat_khan_yang_shao_2020} was reported by multiple participants.
    }
    \label{tab:bench}
    \vspace{2mm}
    \resizebox{\linewidth}{!}{
    \begin{tabular}{l||c|c||c|c||c|c||c|c}
        \hline\noalign{\smallskip}
        & \multicolumn{4}{c ||}{\textbf{Track 1 (Samsung S7)}} & \multicolumn{4}{c}{\textbf{Track 2 (Huawei P20)}} \\
        Team & \multicolumn{2}{c||}{Test1} & \multicolumn{2}{c||}{Test2} & \multicolumn{2}{c||}{Test1} & \multicolumn{2}{c}{Test2} \\
         name & PSNR~$\uparrow$ & SSIM~$\uparrow$ & PSNR~$\uparrow$ & SSIM~$\uparrow$ & PSNR~$\uparrow$ & SSIM~$\uparrow$ & PSNR~$\uparrow$ & SSIM~$\uparrow$ \\
        \hline
        \rowcolor{Gray} NOAHTCV	 & 31.86 & 0.83 & 32.69 & 0.88 & 38.38 & 0.93 & 35.77 & 0.92 \\
        MiAlgo	    & 31.39 & 0.82 & 30.73 & 0.80 & 40.06 & 0.93 & 37.09 & 0.92  \\
        \rowcolor{Gray} CASIA LCVG (*) & 30.19 & 0.81 & 31.47 & 0.86 & 37.58 & 0.93 & 33.99 & 0.92  \\
        HIT-IIL	    & 29.12 & 0.80 & 30.22 & 0.87  & 36.53 & 0.91 & 34.07 & 0.90 \\
        \rowcolor{Gray} SenseBrains (Ours)	& 28.36 & 0.80 & 30.08 & 0.86 & 35.47 & 0.92 & 32.63 & 0.91 \\
        CS\textasciicircum2U (*)  & 29.13 & 0.79 & 29.95 & 0.84 & - & - & - & -\\
        \rowcolor{Gray} HiImage	    & 27.96 & 0.79 & - & - & 34.40 & 0.94 & 32.13 & 0.90 \\
        0noise	    & 27.67 & 0.79 & 29.81 & 0.87 & 33.68 & 0.90 & 31.83 & 0.89 \\
        \rowcolor{Gray} OzU VGL	    & 27.89 & 0.79 & 28.83 & 0.83 & 32.72 & 0.87 & 30.69 &0.86 \\
        PixelJump   & 28.15 & 0.80 & - & - & -  & - & - & - \\
        \rowcolor{Gray} CVIP & 27.85 & 0.80 & 29.50 & 0.86 & -  & - & - & - \\
        \hline
        CycleISP~\cite{zamir_arora_khan_hayat_khan_yang_shao_2020}  & 26.75 & 0.78 & - & - & 32.70 & 0.85 & - & -  \\
        UPI~\cite{brooks_mildenhall_xue_chen_sharlet_barron_2019}  & 26.90 & 0.78 & - & - & - & - & - & -  \\
        U-Net Base & 26.30 & 0.77 & - & - & 30.01 & 0.80 & - & -  \\
        \hline
    \end{tabular}}
\end{table}

\begin{table}[!hb]
    \centering
    \caption{Team information summary. Input refers to the input image size used during training, most teams used the provided patches (504px or 496px). ED indicates the use of Extra Datasets besides the provided challenge datasets. ENS indicates if the solution is an Ensemble of multiple models. FR indicates if the model can process Full-Resolution images ($3024\times4032$)}
    \vspace{2mm}
    \label{tab:teams}
    \resizebox{\linewidth}{!}{
    \begin{tabular}{l|c|c|c|c|c|c|c|c}
        \hline
        Team & Input & Epochs & ED & ENS & FR  & \# Params. (M) & Runtime (ms) & GPU \\
        \hline
        \rowcolor{Gray} NOAHTCV & (504,504) & 500 & \xmark & \xmark & \cmark & 5.6 & 25 & V100 \\
        MiAlgo & (3024,4032) & 3000 & \xmark & \xmark & \cmark & 4.5 & 18 & V100 \\
        \rowcolor{Gray} CASIA LCVG & (504,504) & 300K it. & \cmark & \cmark & \cmark & 464 & 219 & A100 \\
        CS\textasciicircum2U & (504,504) & 276K it. & \cmark & \cmark & \cmark & 105 & 1300 & 3090 \\
        \rowcolor{Gray} HIT-IIL & (1536,1536) & 1000 & \xmark & \xmark & \cmark & 116 & 19818 & V100 \\
        SenseBrains & (504,504) & 220 & \xmark & \cmark & \cmark & 69 & 50 & V100 \\
        \rowcolor{Gray} PixelJump & (504,504) & 400  & \xmark & \cmark & \cmark & 6.64 & 40 & 3090 \\
        HiImage & (256, 256) & 600 & \xmark & \xmark & \cmark & 11 & 200 & 3090 \\
        \rowcolor{Gray} OzU VGL & (496, 496) & 52 & \xmark & \xmark & \cmark & 86 & 6 & 2080 \\
        CVIP & (504,504) & 75 & \xmark & \xmark & \cmark & 2.8 & 400 & 3090 \\
        \rowcolor{Gray} 0noise & (504,504) & 200 & \xmark & \xmark & \cmark & 0.17 & 19 & Q6000 \\
        \hline
    \end{tabular}
    }
\end{table}

Our proposed methodology (SenseBrains) along with other solutions were quantitatively evaluated in fidelity measures as well as generalizability and robustness. As listed in Table~\ref{tab:bench}, our pipeline improves the performance of the U-Net from $26.30$ dB to $28.36$ dB in Track 1 and from $30.01$ dB to $35.47$ dB in Track 2 . Our methodology uniformly fixed all neural networks used to be the standard U-Net \cite{DBLP:journals/corr/RonnebergerFB15} with $23$ convolutional layers with only an additional channel for the overexposure mask, which subtantiates the enhancement in performance through the multi-step refinement process. 

Table~\ref{tab:teams} provides further details on training as well as additional details such as use of extra data, ensemble or whether it is capable of processing full-resolution images. Note that although our methodology using the two U-Nets have a relatively high number of model parameters, by replacing the U-Net with other methodologies that supports end-to-end learning, the number of model parameters along with runtime can be controlled. 

\begin{figure}[!ht] 
        \centering 
        \begin{subfigure}{0.495\textwidth} 
            \centering
            \includegraphics[width=\textwidth]{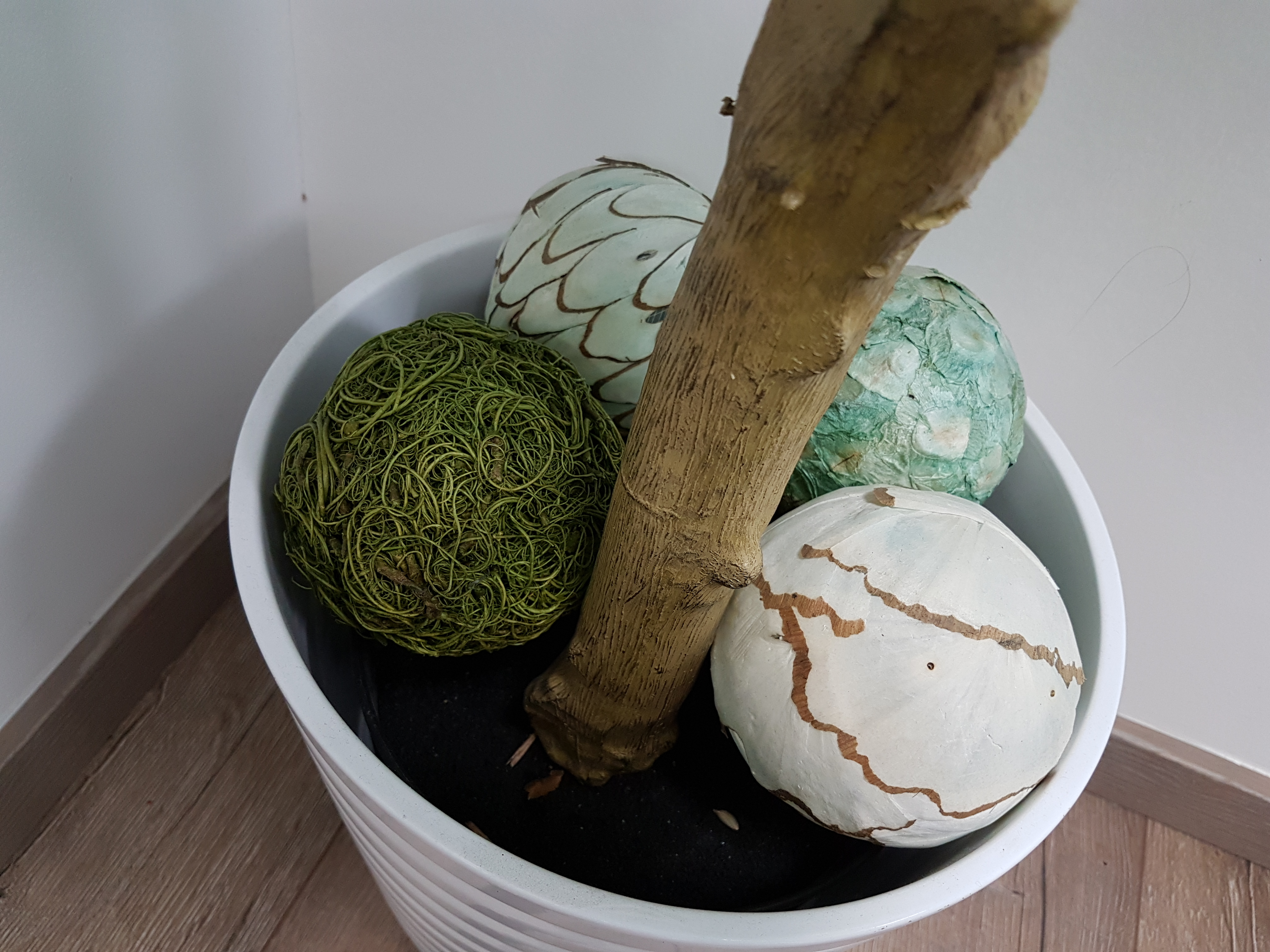} 
            \caption{Input Full Resolution Image}
            \label{subfig:full_resolution_original} 
        \end{subfigure} 
        \hfill
        \begin{subfigure}{0.495\textwidth} 
            \centering
            \includegraphics[width=\textwidth]{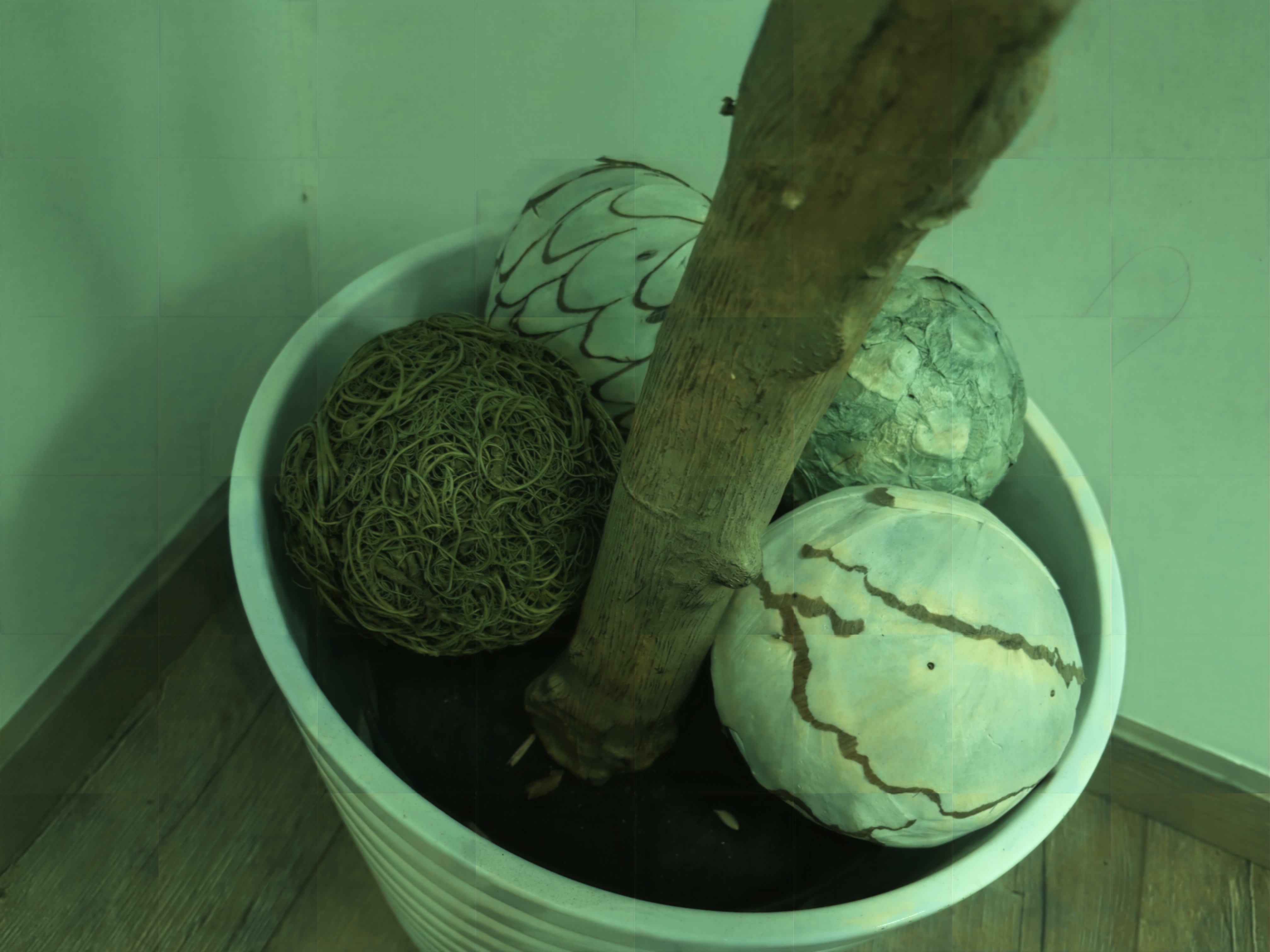}
            \caption{Visualized Output}
            \label{subfig:full_resolution_original_2}
        \end{subfigure} 
    \caption{(a) a full resolution image ($3024 \times 4032$) (b) visualized RAW output from our pipeline. Visualization of RAW was performed by passing the output RAW through  Demosaic Net \cite{gharbi_chaurasia_paris_durand_2016} along with a simple implementation of post-processing. Note that our pipeline is able to perform inference on full resolution images.} 
    \label{full_resolution_compare}
\end{figure} 

\begin{figure}[!ht] 
        \centering 
        \begin{subfigure}{0.155\textwidth} 
            \centering
            \caption{} 
            \includegraphics[width=\textwidth]{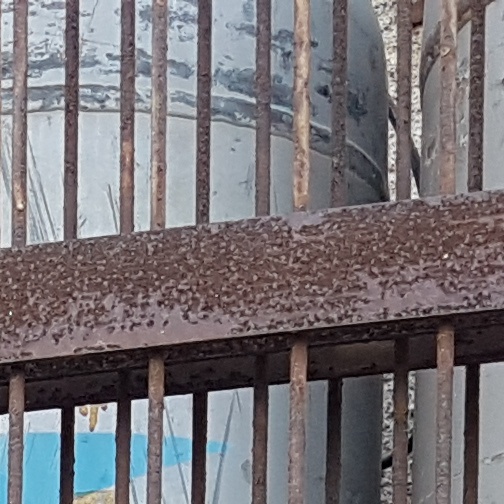} 
            \label{sub:samples_compare_1} 
        \end{subfigure} 
        \hfill
        \begin{subfigure}{0.155\textwidth} 
            \centering
            \caption{}
            \includegraphics[width=\textwidth]{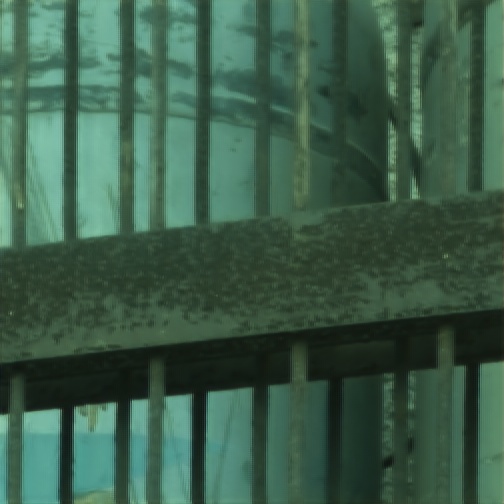} 
            \label{sub:samples_compare_2} 
        \end{subfigure} 
        \hfill
        \begin{subfigure}{0.155\textwidth} 
            \centering
            \caption{}
            \includegraphics[width=\textwidth]{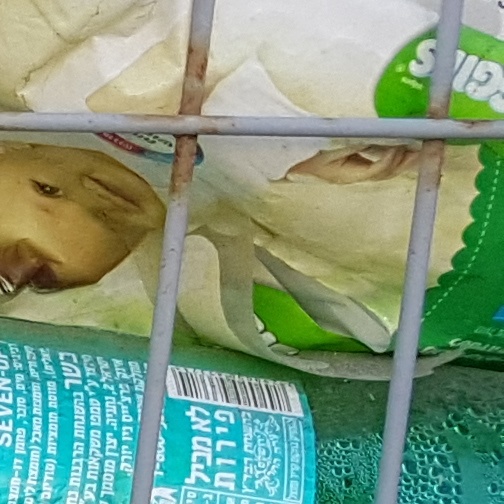} 
            \label{sub:samples_compare_3} 
        \end{subfigure} 
        \hfill
        \begin{subfigure}{0.155\textwidth} 
            \centering
            \caption{}
            \includegraphics[width=\textwidth]{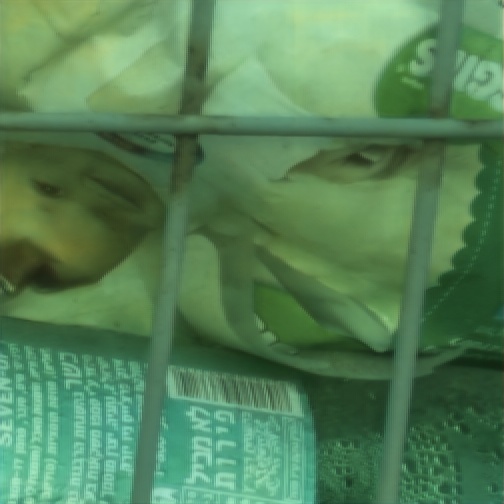} 
            \label{sub:samples_compare_4} 
        \end{subfigure} 
        \hfill
        \begin{subfigure}{0.155\textwidth} 
            \centering
            \caption{}
            \includegraphics[width=\textwidth]{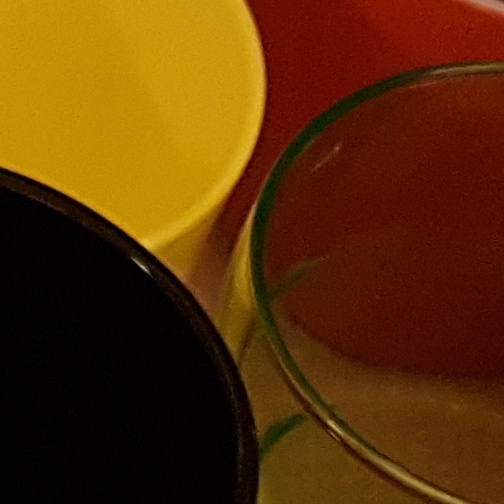} 
            \label{sub:samples_compare_5} 
        \end{subfigure} 
        \hfill
        \begin{subfigure}{0.155\textwidth} 
            \centering
            \caption{} 
            \includegraphics[width=\textwidth]{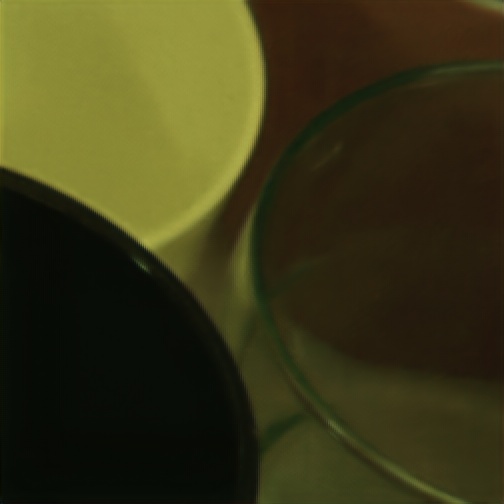} 
            \label{sub:samples_compare_6} 
        \end{subfigure} 
        \hfill
        \begin{subfigure}{0.155\textwidth} 
            \centering
            \includegraphics[width=\textwidth]{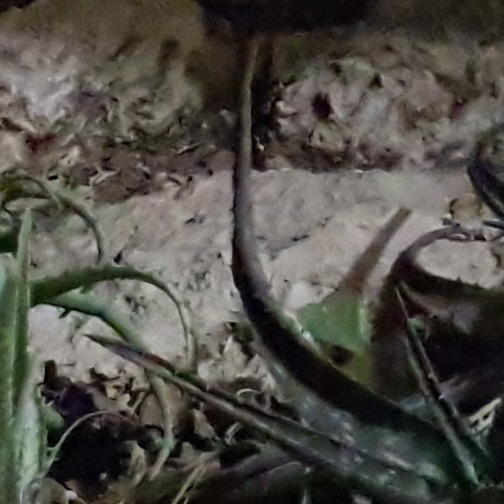} 
            \label{sub:samples_compare_7} 
        \end{subfigure} 
        \hfill
        \begin{subfigure}{0.155\textwidth} 
            \centering
            \includegraphics[width=\textwidth]{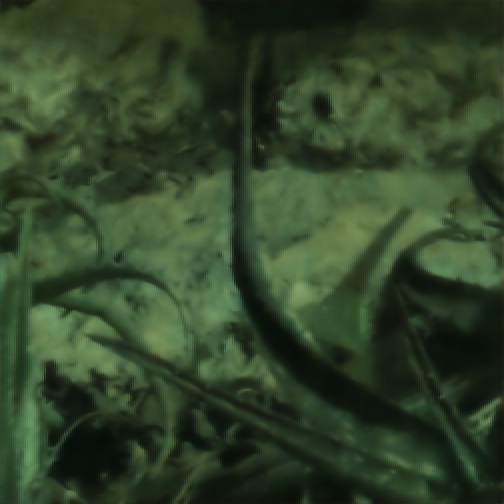} 
            \label{sub:samples_compare_8} 
        \end{subfigure} 
        \hfill
        \begin{subfigure}{0.155\textwidth} 
            \centering
            \includegraphics[width=\textwidth]{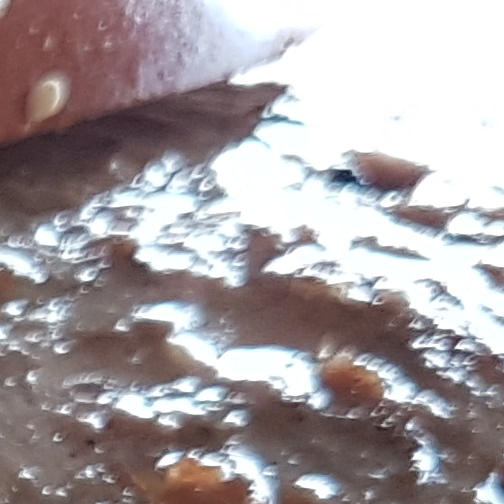} 
            \label{sub:samples_compare_9} 
        \end{subfigure} 
        \hfill
        \begin{subfigure}{0.155\textwidth} 
            \centering
            \includegraphics[width=\textwidth]{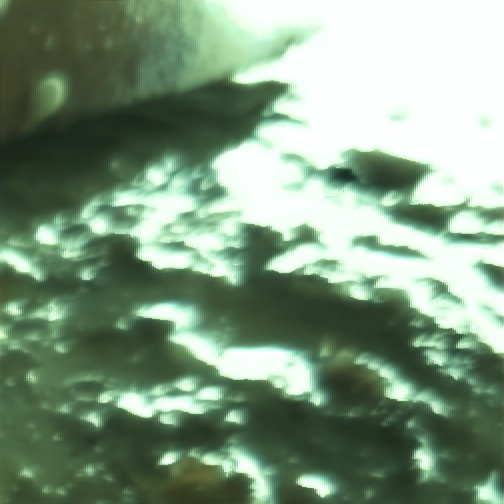} 
            \label{sub:samples_compare_10} 
        \end{subfigure} 
        \hfill
        \begin{subfigure}{0.155\textwidth} 
            \centering
            \includegraphics[width=\textwidth]{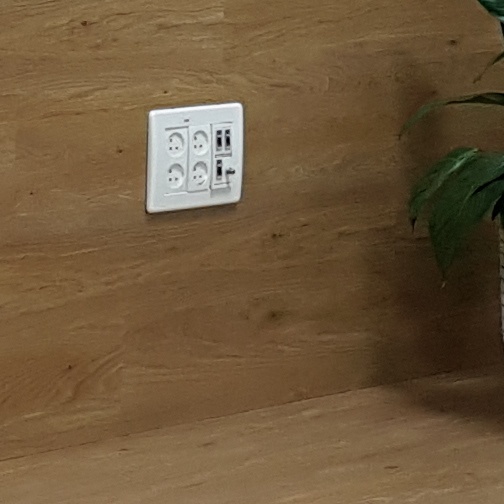} 
            \label{sub:samples_compare_11} 
        \end{subfigure} 
        \hfill
        \begin{subfigure}{0.155\textwidth} 
            \centering
            \includegraphics[width=\textwidth]{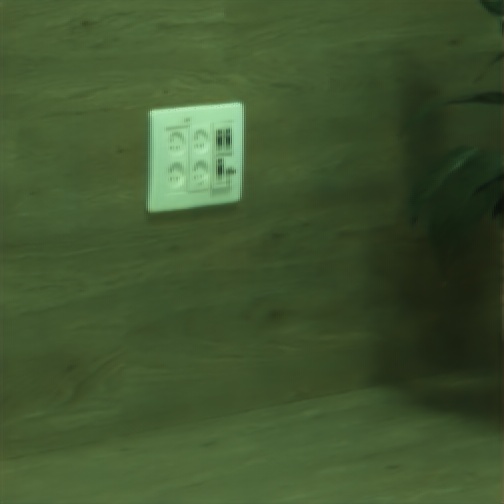} 
            \label{sub:samples_compare_12} 
        \end{subfigure} 
        \hfill
        \begin{subfigure}{0.155\textwidth} 
            \centering
            \includegraphics[width=\textwidth]{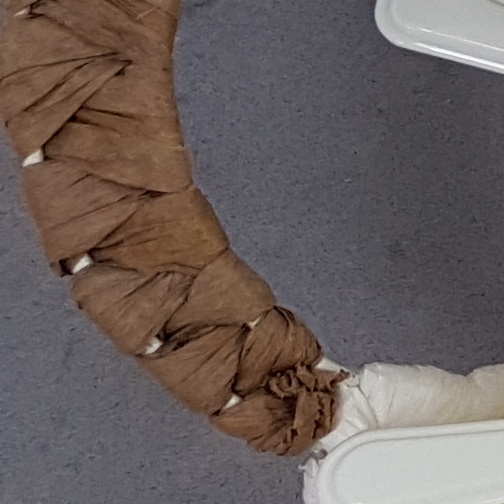} 
            \label{sub:samples_compare_80} 
        \end{subfigure} 
        \hfill
        \begin{subfigure}{0.155\textwidth} 
            \centering
            \includegraphics[width=\textwidth]{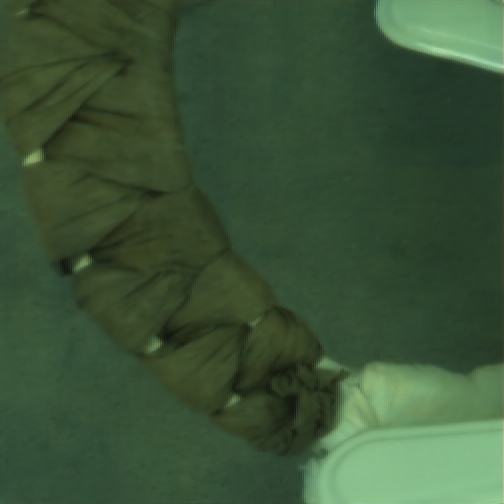} 
            \label{sub:samples_compare_90} 
        \end{subfigure} 
        \hfill
        \begin{subfigure}{0.155\textwidth} 
            \centering
            \includegraphics[width=\textwidth]{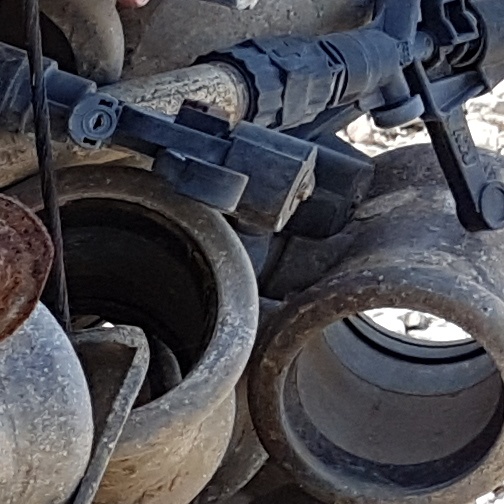} 
            \label{sub:samples_compare_100} 
        \end{subfigure} 
        \hfill
        \begin{subfigure}{0.155\textwidth} 
            \centering
            \includegraphics[width=\textwidth]{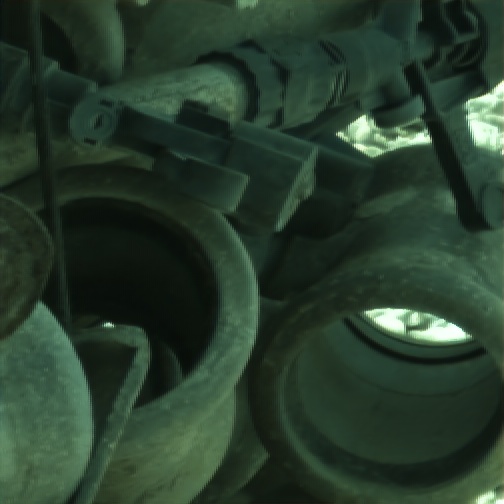} 
            \label{sub:samples_compare_112} 
        \end{subfigure} 
        \hfill
        \begin{subfigure}{0.155\textwidth} 
            \centering
            \includegraphics[width=\textwidth]{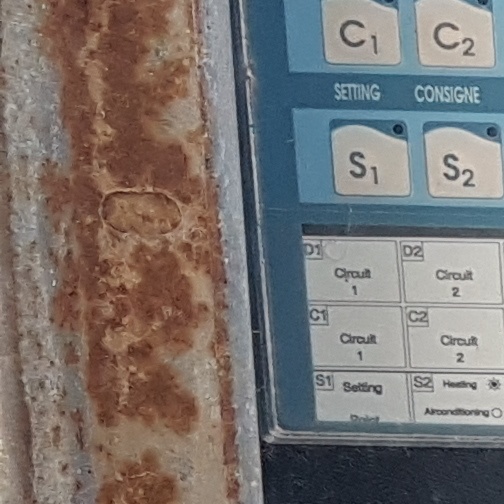} 
            \label{sub:samples_compare_128} 
        \end{subfigure} 
        \hfill
        \begin{subfigure}{0.155\textwidth} 
            \centering
            \includegraphics[width=\textwidth]{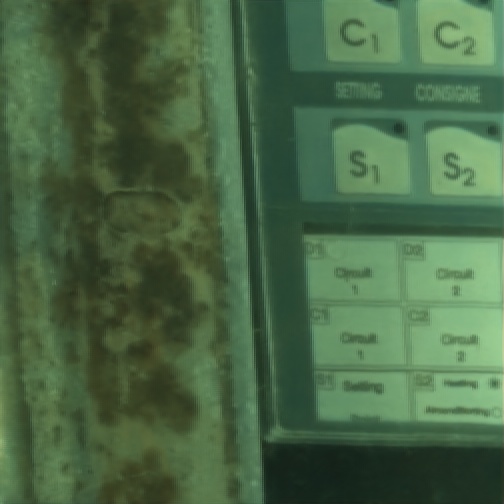} 
            \label{sub:samples_compare_130} 
        \end{subfigure} 
    \caption{The figure displays input RGB images under columns (a), (c) and (e) and corresponding visualized RAW outputs under columns (b), (d) and (f). Note that the displayed RGB images have a variety of colors and textures. Visualization of RAW was performed by passing RAW outputs into Demosaic Net \cite{gharbi_chaurasia_paris_durand_2016} and then a simplified implementation of post-processing.} 
    \label{trained_images_compare}
\end{figure}

\begin{figure}[!ht] 
        \centering 
        \begin{subfigure}{0.155\textwidth} 
            \centering
            \caption{}
            \includegraphics[width=\textwidth]{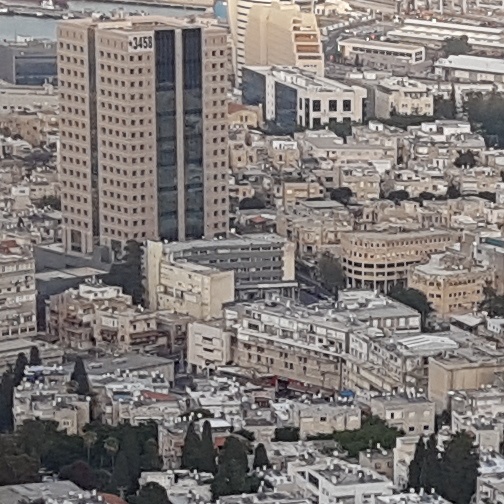} 
            \label{subfig:team_samples_compare_1} 
        \end{subfigure} 
        \hfill
        \begin{subfigure}{0.155\textwidth} 
            \caption{}
            \includegraphics[width=\textwidth]{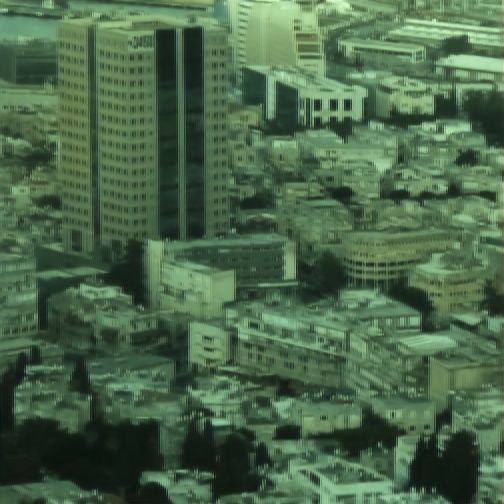} 
            \label{subfig:team_samples_compare_2} 
        \end{subfigure} 
        \hfill
        \begin{subfigure}{0.155\textwidth} 
            \caption{} 
            \includegraphics[width=\textwidth]{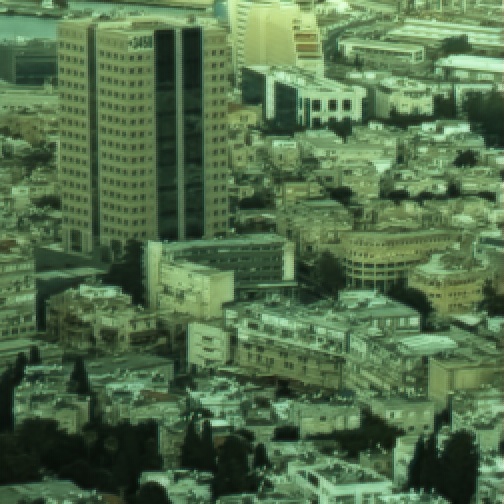} 
            \label{subfig:team_samples_compare_3} 
        \end{subfigure} 
        \hfill
        \begin{subfigure}{0.155\textwidth} 
            \caption{} 
            \includegraphics[width=\textwidth]{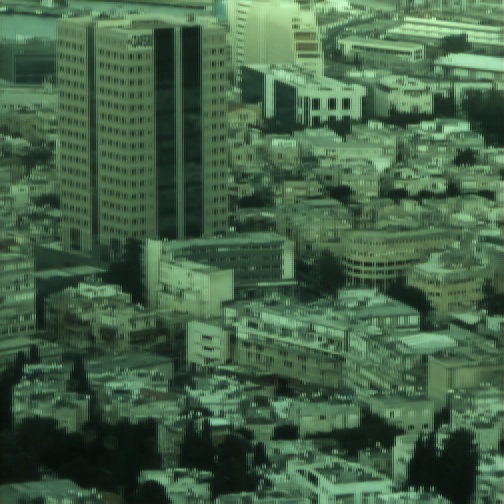} 
            \label{subfig:team_samples_compare_4} 
        \end{subfigure} 
        \hfill
        \begin{subfigure}{0.155\textwidth} 
            \caption{} 
            \includegraphics[width=\textwidth]{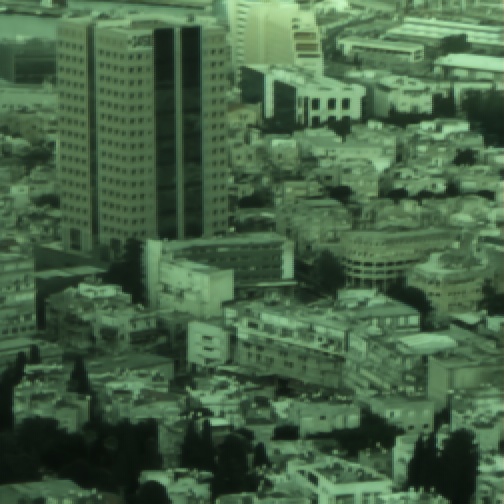} 
            \label{subfig:team_samples_compare_5} 
        \end{subfigure} 
        \hfill
        \begin{subfigure}{0.155\textwidth} 
            \caption{} 
            \includegraphics[width=\textwidth]{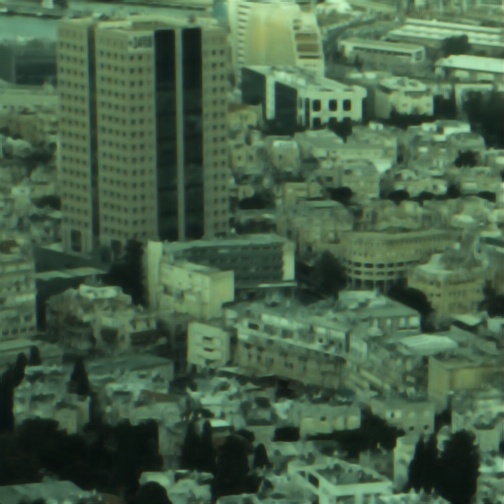} 
            \label{subfig:team_samples_compare_6} 
        \end{subfigure} 
        \hfill
        \begin{subfigure}{0.155\textwidth} 
            \caption{} 
            \includegraphics[width=\textwidth]{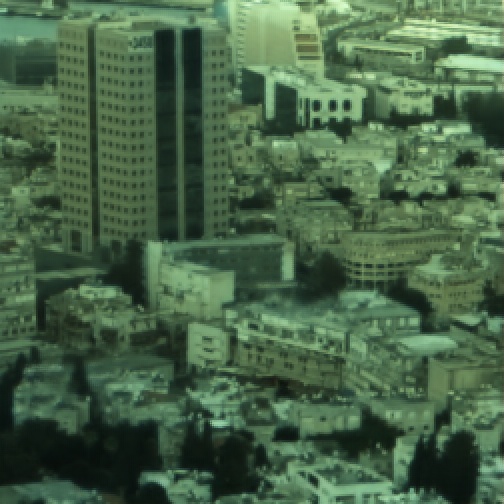} 
            \label{subfig:team_samples_compare_7} 
        \end{subfigure} 
        \hfill
        \begin{subfigure}{0.155\textwidth} 
            \caption{} 
            \includegraphics[width=\textwidth]{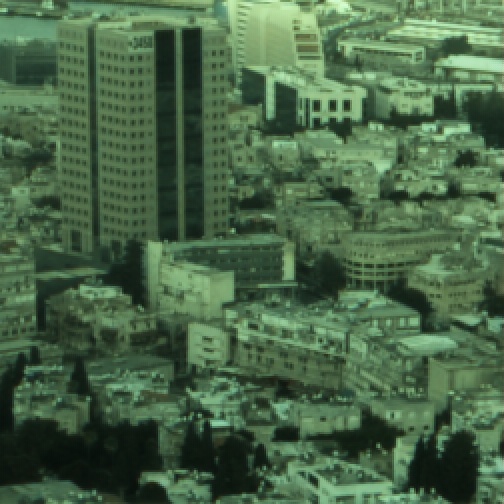} 
            \label{subfig:team_samples_compare_8} 
        \end{subfigure} 
        \hfill
        \begin{subfigure}{0.155\textwidth} 
        \caption{}
            \includegraphics[width=\textwidth]{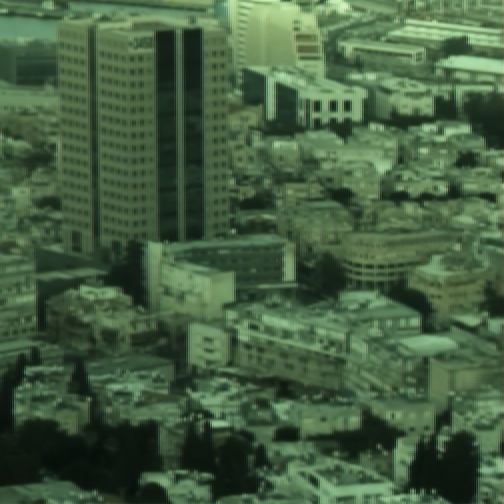} 
            \label{subfig:team_samples_compare_9} 
        \end{subfigure} 
        \hfill
        \begin{subfigure}{0.155\textwidth} 
            \caption{} 
            \includegraphics[width=\textwidth]{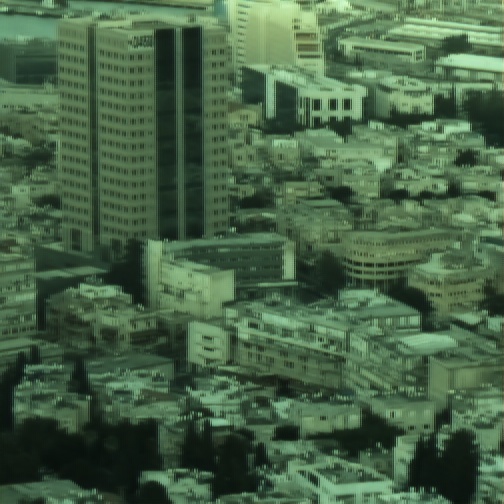} 
            \label{subfig:team_samples_compare_10} 
        \end{subfigure} 
        \hfill
        \begin{subfigure}{0.155\textwidth} 
            \caption{} 
            \includegraphics[width=\textwidth]{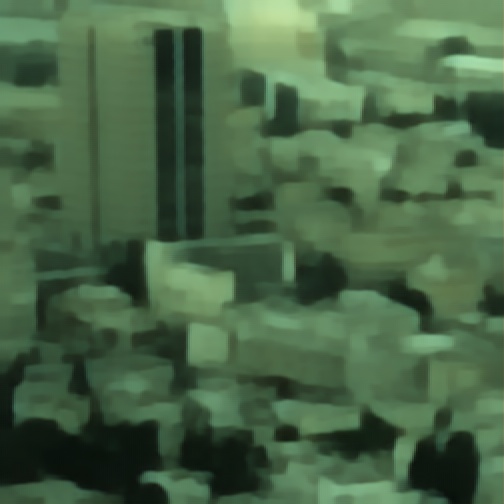} 
            \label{subfig:team_samples_compare_11} 
        \end{subfigure} 
        \hfill
        \begin{subfigure}{0.155\textwidth} 
            \caption{} 
            \includegraphics[width=\textwidth]{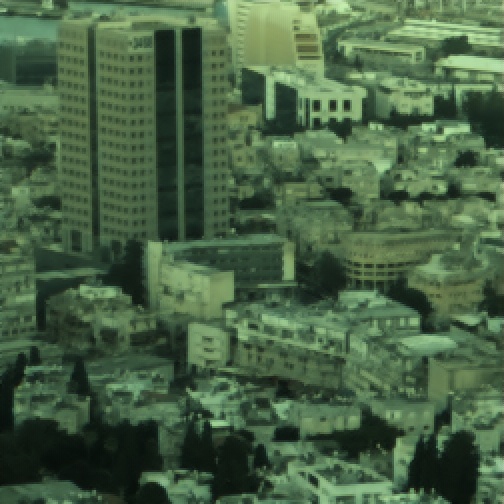}
            \label{subfig:team_samples_compare_12} 
        \end{subfigure} 
    \caption{(a) is the input RGB image. From (b) to (l), the outputs of the following teams are displayed in the following order: SenseBrains, 0noise, CASIA LCVG, CS\textasciicircum2U, CVIP, HiImage, HIT-IIL, MiAlgo, NOAHTCV, OzU VGL, Pixel Jump. All RAW outputs were visualized by passing into Demosaic Net \cite{gharbi_chaurasia_paris_durand_2016} and then a simplified implementation of post-processing.} 
    \label{team_samples_compare}
\end{figure}

In Fig.~\ref{full_resolution_compare}, an input full resolution image ($3024 \times 4032$), as a collection of patches is passed into our pipeline. The RAW output was generated by constructing the overexposure mask, passing the RGB image into the two U-Nets, and mosaicing the blended output. The RAW output was then visualized with Demosaic Net to acquire the demosaiced RAW and then passed into a simplified implementation of post-processing. It is noteworthy that our pipeline with U-Net (OE) and U-Net (NOE) can perform inference on full resolution images and produce visualized RAW images that show high perceptual quality, as shown in Fig.~\ref{full_resolution_compare}. Additionally, Fig.~\ref{trained_images_compare} displays corresponding pairs of RGB images and visualized RAW outputs produced by our pipeline. Note that the RGB images have a variety of colors and textures, including images also with overexposure for which using the overexposure mask fusion process resulted in higher fidelity performance. Furthermore, Fig.~\ref{team_samples_compare} displays a comparison of visualized RAW outputs of all proposed methodologies. 

\subsection{Ablation Studies} 

By experimenting with the training and validation dataset for the Samsung S7 dataset \cite{schwartz_giryes}, we performed ablation studies along with the quantitative evaluations provided by the AIM Reversed ISP Challenge \cite{conde2022aim}. Note that these datasets are public, and are only used for ablation studies and to evaluate refinement processes of our proposed pipeline. 

\setlength{\tabcolsep}{4pt} 
\begin{table}
    \centering
    \caption{Samsung S7 Training Dataset \cite{schwartz_giryes} Comparison. The comparison was conducted for ablation purposes and to evaluate steps of the proposed pipeline. Evaluation was conducted by computing the average PSNR over the training dataset.}
        \label{abelation_table_1}
        \begin{tabular}{c|c|c}
            \hline\noalign{\smallskip} 
            Model & Total Number of params (M) & Training PSNR (dB) \\ 
            \hline 
            \rowcolor{Gray} U-Net (OE), U-Net (NOE) & 69.05 & 28.9\\
            U-Net (OE) & 34.5 & 27.23\\
            \rowcolor{Gray} U-Net \cite{DBLP:journals/corr/RonnebergerFB15} & 34.5 & 26.295\\
            \hline
        \end{tabular}
    \end{table}
\setlength{\tabcolsep}{1.4pt} 

As recorded in Table ~\ref{abelation_table_1}, the baseline U-Net mapping directly from RGB to RAW bayer has an average training PSNR of $26.295$ dB. One single U-Net with the addition of a channel for the overexposure mask and trained to map from RGB to demosaiced RAW has a average training PSNR of $27.23$ dB. Despite the only significant modification being the addition of a channel to pass the overexposure mask, mapping from RGB to demosaiced RAW leads to improvement in fidelity. Additionally with the blending of two separate neural networks for overexposed and non-overexposed pixels, the training PSNR increases to $28.9$ dB. 

\setlength{\tabcolsep}{4pt}
\begin{table}
    \centering
    \caption{Samsung S7 Validation Dataset PSNR \cite{schwartz_giryes} Comparison. The comparison was conducted for ablation purposes and to evaluate steps of the proposed pipeline. Evaluation was conducted on the validation dataset.}
        \label{abelation_table_2}
        \begin{tabular}{c|c|c}
            \hline\noalign{\smallskip}
            Model & Total Number of params (M) & Validation PSNR (dB) \\ 
            \hline 
            \rowcolor{Gray} U-Net (OE) & 34.5 & 32.17\\
            UPI \cite{brooks_mildenhall_xue_chen_sharlet_barron_2019}, YUV refinement & 34.5 & 29.9\\
            \rowcolor{Gray} UPI \cite{brooks_mildenhall_xue_chen_sharlet_barron_2019}& - & 26.97\\
            \hline
        \end{tabular}
    \end{table}
\setlength{\tabcolsep}{1.4pt}

Another comparison that was made on the Samsung S7 validation dataset was with UPI \cite{brooks_mildenhall_xue_chen_sharlet_barron_2019} and the additional step of YUV refinement offered in the pipeline as displayed in Fig.~\ref{network_architecture}. Note that UPI \cite{brooks_mildenhall_xue_chen_sharlet_barron_2019} had a validation PSNR of $26.97$ dB, while YUV refinement improves fidelity by $2.93$ dB. Even with this refinement, however, U-Net (OE), which is a single U-Net mapping from RGB images to demosaiced RAW, has a significantly higher validation PSNR of $32.17$ dB.

\begin{figure}[!ht] 
        \centering 
        \begin{subfigure}{0.155\textwidth} 
            \centering
            \caption{}
            \includegraphics[width=\textwidth]{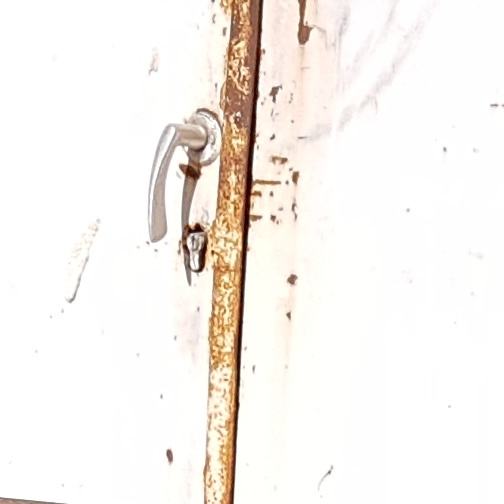} 
            \label{subfig:samples_comp_1} 
        \end{subfigure} 
        \hfill
        \begin{subfigure}{0.155\textwidth} 
            \centering
            \caption{}
            \includegraphics[width=\textwidth]{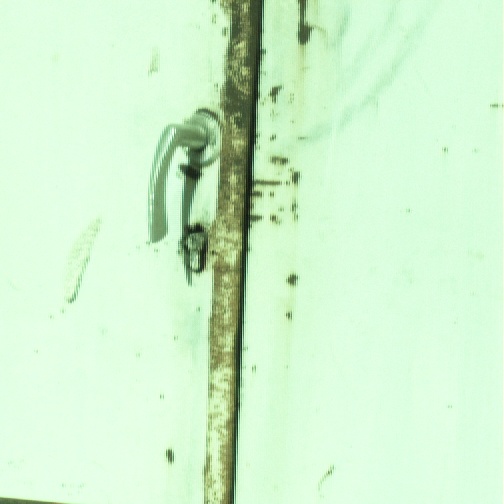} 
            \label{subfig:samples_comp_2} 
        \end{subfigure} 
        \hfill 
        \begin{subfigure}{0.155\textwidth} 
            \centering
            \caption{}
            \includegraphics[width=\textwidth]{paper_image/trained_images/39_14.jpg} 
            \label{subfig:samples_comp_100} 
        \end{subfigure} 
        \hfill
        \begin{subfigure}{0.155\textwidth} 
            \centering
            \caption{}
            \includegraphics[width=\textwidth]{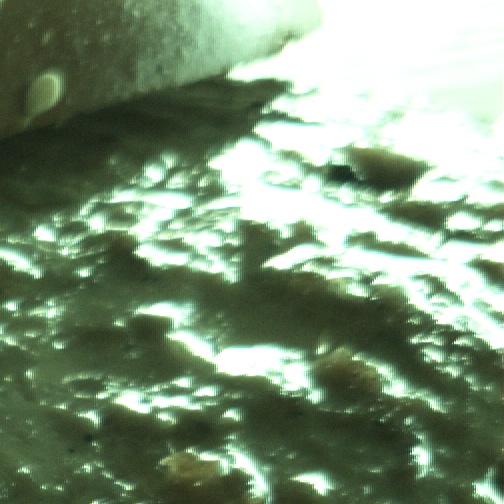} 
            \label{subfig:samples_comp_200} 
        \end{subfigure} 
        \hfill 
        \begin{subfigure}{0.155\textwidth} 
            \centering
            \caption{}
            \includegraphics[width=\textwidth]{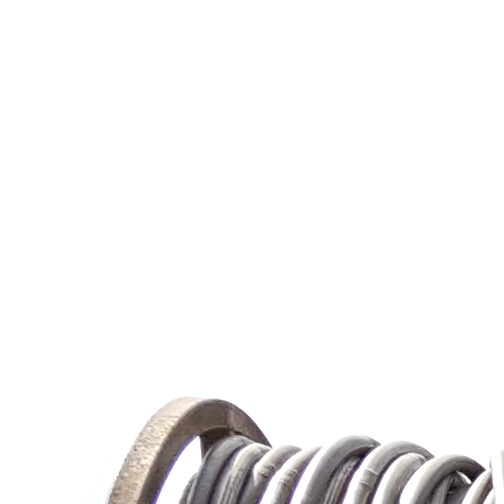} 
            \label{subfig:samples_comp_3} 
        \end{subfigure} 
        \hfill
        \begin{subfigure}{0.155\textwidth} 
            \centering
            \caption{}
            \includegraphics[width=\textwidth]{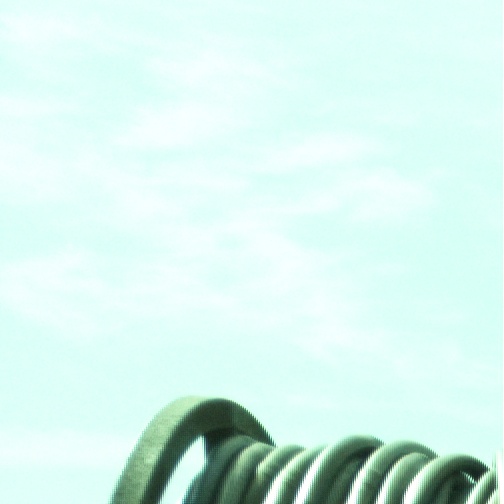} 
            \label{subfig:samples_comp_4} 
        \end{subfigure}
        \hfill 
        \begin{subfigure}{0.155\textwidth} 
            \centering
            \includegraphics[width=\textwidth]{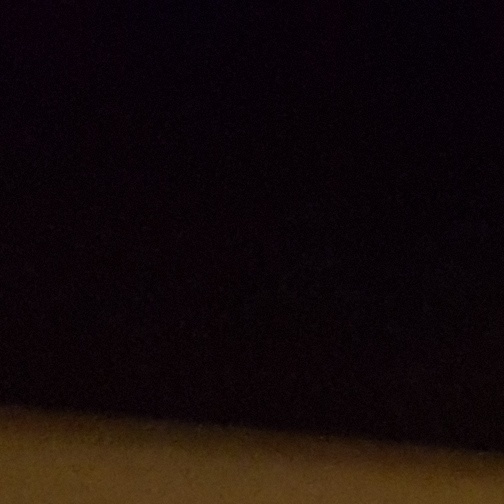} 
            \label{subfig:samples_comp_5} 
        \end{subfigure} 
        \hfill
        \begin{subfigure}{0.155\textwidth} 
            \centering
            \includegraphics[width=\textwidth]{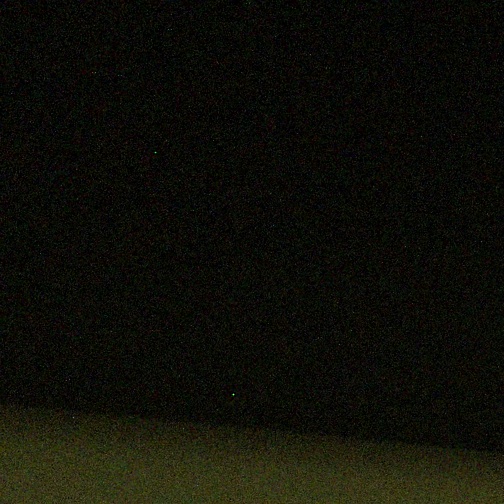} 
            \label{subfig:samples_comp_6} 
        \end{subfigure}
        \hfill 
        \begin{subfigure}{0.155\textwidth} 
            \centering
            \includegraphics[width=\textwidth]{paper_image/trained_images/82_28.jpg} 
            \label{subfig:samples_comp_7} 
        \end{subfigure} 
        \hfill
        \begin{subfigure}{0.155\textwidth} 
            \centering
            \includegraphics[width=\textwidth]{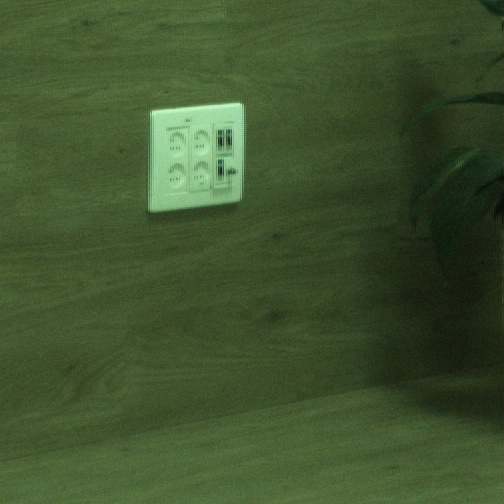} 
            \label{subfig:samples_comp_8} 
        \end{subfigure}
        \hfill 
        \begin{subfigure}{0.155\textwidth} 
            \centering
            \includegraphics[width=\textwidth]{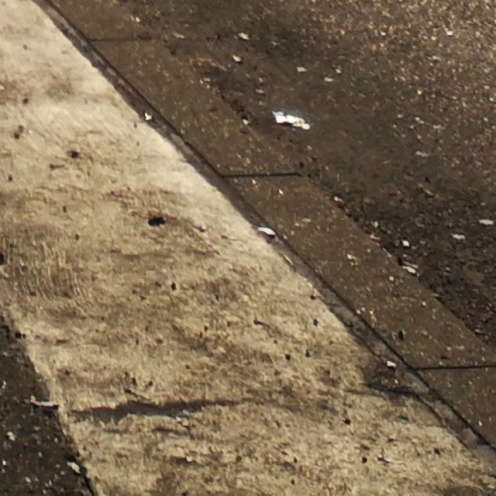} 
            \label{subfig:samples_comp_9} 
        \end{subfigure} 
        \hfill
        \begin{subfigure}{0.155\textwidth} 
            \centering
            \includegraphics[width=\textwidth]{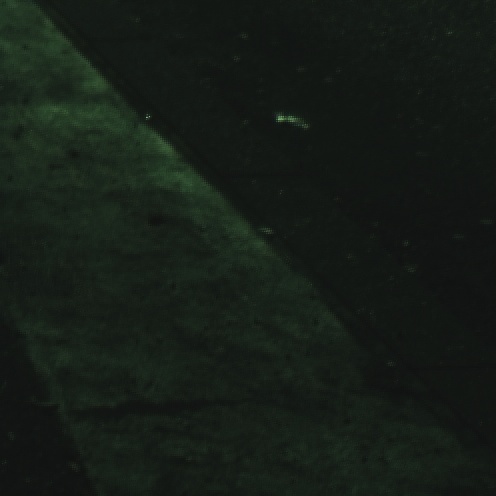} 
            \label{subfig:samples_comp_10} 
        \end{subfigure}
    \caption{The figure displays input RGB images under columns (a), (c) and (e) and the corresponding visualized RAW groundtruth images under columns (b), (d) and (f). Note that the input RGB images have low-light demosaiced RAW or overexposure. Visualization of RAW was performed by passing RAW outputs into Demosaic Net  \cite{gharbi_chaurasia_paris_durand_2016} and then a simplified implementation of post-processing.} 
    \label{samples_images}
\end{figure}

It is noteworthy that methodologies without use of the overexposure mask fusion process had lowest performance in PSNR for images with overexposure as well as images corresponding to low-light demosaiced RAW images as shown in Fig.~\ref{samples_images}. After visualizing groundtruth bayer using only Demosaic Net \cite{gharbi_chaurasia_paris_durand_2016} without post-processing, we observed that the same white overexposed pixels map  to white overexposed pixels as well as green white overexposed pixels in demosaiced RAW, resulting in drops in fidelity measures for models that do not account for such contradictions. The proposed overexposure mask fusion better accounts for overexposed pixels by designating a U-Net (OE) to overexposed pixels. 

\subsection{Limitations} 
As noted by Conde et al. \cite{conde_mcdonagh_maggioni_leonardis_pérez-pellitero_2022}, one major difficulty that arises with the task of the reversed ISP is the issue of overexposure in input RGB images. This paper addresses overexposure with the proposed methodology of overexposure mask fusion.

A limitation of our pipeline is that mapping from RGB images to demosaiced RAW requires the use of methodologies of reconstructing demosaiced RAW from the groundtruth RAW. For our pipeline, we utilize Demosaic Net \cite{gharbi_chaurasia_paris_durand_2016}, however it is possible that inaccuracies in the reconstruction process of the demosaiced RAW will result in inaccuracies from mapping from RGB images to demosaiced RAW. This possibility of generating inaccurate demosaiced RAW exists and can be further addressed by utilizing other, reliable methods. 

Otherwise, there are few limitations to the pipeline itself, as different steps of the refinement process can be replaced and improved such as the U-Net (OE) and U-Net (NOE). Our pipeline can integrate methodologies that support end-to-end learning and capable of modification to map from RGB to demosaiced RAW. Other issues such as misalignment of RAW and RGB training pairs can be resolved by downsampling before computing perceptual loss. 

\section{Conclusion} 

In this paper, we propose a novel and generalizable multi-step pipeline that allows the use of perceptual loss by mapping from RGB images to demosaiced RAW. With overexposure mask fusion to address overexposure in input RGB images and gamma correction before computing loss for non-overexposed pixels, our methodology addresses several major complexities that have existed for the task of reversed ISP. With significant improvement using only the U-Net \cite{DBLP:journals/corr/RonnebergerFB15} to map from RGB to demosaiced RAW, we have created a multi-step refinement process for enhancement in performance of other solutions that are capable of end-to-end learning and modifiable to map from RGB to demosaiced RAW. By proposing a generalizable process of refinement with steps that can be easily replaced and improved upon, our proposed methodology has notable potential to enhance current and future solutions for the task of RAW reconstruction. 

%
%
\bibliographystyle{splncs04} 
\bibliography{egbib}  
\end{document}